\pdfoutput=1

\documentclass[11pt]{article}

\usepackage[preprint]{acl}
\usepackage{tablefootnote}
\usepackage{times}
\usepackage{latexsym}
\usepackage{threeparttable}
\usepackage{amsmath}

\usepackage{tcolorbox}
\usepackage[T1]{fontenc}
\usepackage{soul} 

\usepackage{color, xcolor}

\usepackage[utf8]{inputenc}
\usepackage{enumitem}
\setenumerate[1]{itemsep=1.5pt,partopsep=0pt,parsep=\parskip,topsep=5pt}
\setitemize[1]{itemsep=1.5pt,partopsep=0pt,parsep=\parskip,topsep=5pt}
\usepackage{microtype}

\usepackage{inconsolata}

\usepackage{graphicx}
\title{HARE: an entity and relation centric evaluation framework for histopathology reports}

\author{
Yunsoo Kim$^{1}$, Michal W. S. Ong$^{1}$, Alex Shavick$^{3}$, Honghan Wu$^{1,2}$, Adam P. Levine$^{1,4}$ \\
$^{1}$University College London, London, UK \\
$^{2}$University of Glasgow, Glasgow, UK \\
$^{3}$Human Technopole, Milan, Italy \\
$^{4}$University College London Hospitals NHS Foundation Trust, London, UK \\
\texttt{yunsoo.kim.23@ucl.ac.uk}
}
\begin{document}

\maketitle
\begin{abstract}
Medical domain automated text generation is an active area of research and development; however, evaluating the clinical quality of generated reports remains a challenge, especially in instances where domain-specific metrics are lacking, e.g. histopathology. We propose \textbf{HARE (Histopathology Automated Report Evaluation)}, a novel entity and relation centric framework, composed of a benchmark dataset, a named entity recognition (NER) model, a relation extraction (RE) model, and a novel metric, which prioritizes clinically relevant content by aligning critical histopathology entities and relations between reference and generated reports. To develop the HARE benchmark, we annotated 813 de-identified clinical diagnostic histopathology reports and 652 histopathology reports from The Cancer Genome Atlas (TCGA) with domain-specific entities and relations. We fine-tuned GatorTronS, a domain-adapted language model to develop HARE-NER and HARE-RE which achieved the highest overall F1-score (0.915) among the tested models. The proposed HARE metric outperformed traditional metrics including ROUGE and Meteor, as well as radiology metrics such as RadGraph-XL, with the highest correlation and the best regression to expert evaluations (higher than the second best method, GREEN, a large language model based radiology report evaluator, by Pearson $r = 0.168$, Spearman $\rho = 0.161$, Kendall $\tau = 0.123$, $R^2 = 0.176$, $RMSE = 0.018$). We release HARE, datasets, and the models at \url{https://github.com/knowlab/HARE} to foster advancements in histopathology report generation, providing a robust framework for improving the quality of reports.

\end{abstract}

\section{Introduction}
\label{sec:intro}

Medical report generation has become an increasingly active area of research in clinical natural language processing (NLP) with the goal of automating the creation of specialized clinical documents \cite{xu2024overview,liu2025automaticsurvey}. Among various medical domains, radiology has witnessed the earliest and most notable advancements in automated report generation \cite{hyland2023maira,nicolson2023cxrmate,wu2024slava,bannur2024maira2}. This progress is partly attributed to the development of domain-specific evaluation metrics that prioritize clinical correctness \cite{smit2020chexbert,jain2021radgraph,delbrouck2024radgraphxl,zhao2024ratescore}. Unlike general-purpose metrics such as BLEU and ROUGE, these specialized metrics assess the accuracy of radiologically significant entities and findings, thereby offering a more clinically meaningful measure of report quality \cite{lin2004rouge,papineni2002bleu,zhao2024ratescore} and facilitating the development of accurate generative models.

In contrast, the field of histopathology, which involves the microscopic examination of tissue samples to diagnose diseases such as cancer, still relies only on general-purpose lexical metrics for evaluating automatically generated reports \cite{chen2023migen,guo2024histgen,tan2024clinicalpathgen,chen2024wsicaption}. Histopathology reports are semi-structured, terminology-intensive documents that provide detailed microscopic evaluations of tissue samples. Such reports play a crucial role in disease diagnosis and guiding treatment decisions. Histopathology reports encompass multiple sections, including descriptions of anatomical sites, cellular morphology, tumor classification, staging, further analyses (e.g. immunohistochemistry (IHC) markers, special stains, or in situ hybridization (ISH)), and a final diagnosis/conclusion.

\begin{figure}[t]
    \centering
    \includegraphics[trim={0cm 0cm 0cm 0cm},clip,width=0.5\textwidth]{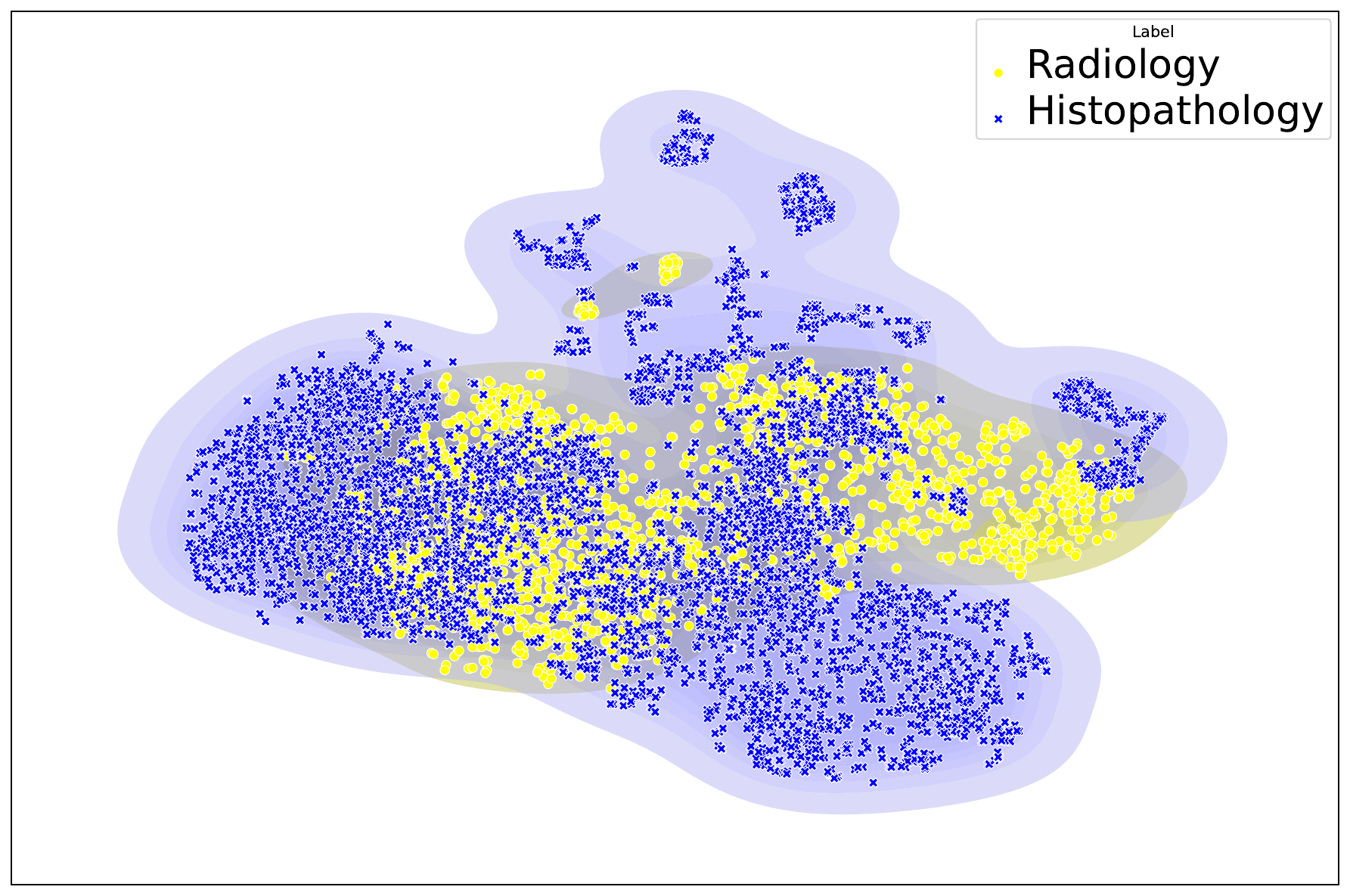}
    \caption{Scatter and density plot of word embeddings for radiology and histopathology reports. The radiology reports are 1,000 randomly sampled reports from MIMIC-CXR dataset and IU-X-ray dataset \cite{johnson2019mimic,Dina2016IU-Xray}. The histopathology reports are 1,000 randomly sampled reports from both datasets used in this study. Reports are embedded using a BERT-base model and reduced to two dimensions using principal component analysis. The density regions highlight where words from each category are concentrated, with "Radiology" shown in yellow and "Histopathology" in blue.}
    \label{fig:scatter_density}
\end{figure}

Figure \ref{fig:scatter_density} shows the difference between the word embeddings of radiology reports (from MIMIC-CXR \cite{johnson2019mimic} and IU-Xray \cite{Dina2016IU-Xray} and histopathology reports (used in this study). Histopathology word embedding has many areas that are not covered by radiology word embeddings, making radiology report evaluation metrics unsuitable for histopathology reports. Conventional lexical evaluation metrics such as METEOR and BERTScore as well as clinical relevance-based evaluation metrics designed for radiology reports are insufficient for assessing the quality of automatically generated histopathology reports, as they fail to capture the nuanced histopathological details essential for accurate diagnosis and patient management \cite{banerjee2005meteor,zhang2019bertscore,smit2020chexbert,delbrouck2024radgraphxl,zhao2024ratescore}.

This challenge is further compounded by the scarcity of publicly available datasets for specifically histopathology named entity recognition (NER) and relation extraction (RE), which limits the ability to train robust models tailored to the complexities of histopathological language. There is only one NER model and dataset for pathology reports; however, these are not publicly available \cite{zeng2023improving}. This gap underscores the need for an entity and relation centric evaluation metric that can capture the unique characteristics of histopathology reports.

\begin{figure*}[htbp]
    \centering
    \fbox{\includegraphics[trim={0cm 0cm 0cm 0cm},clip,width=1.0\textwidth]{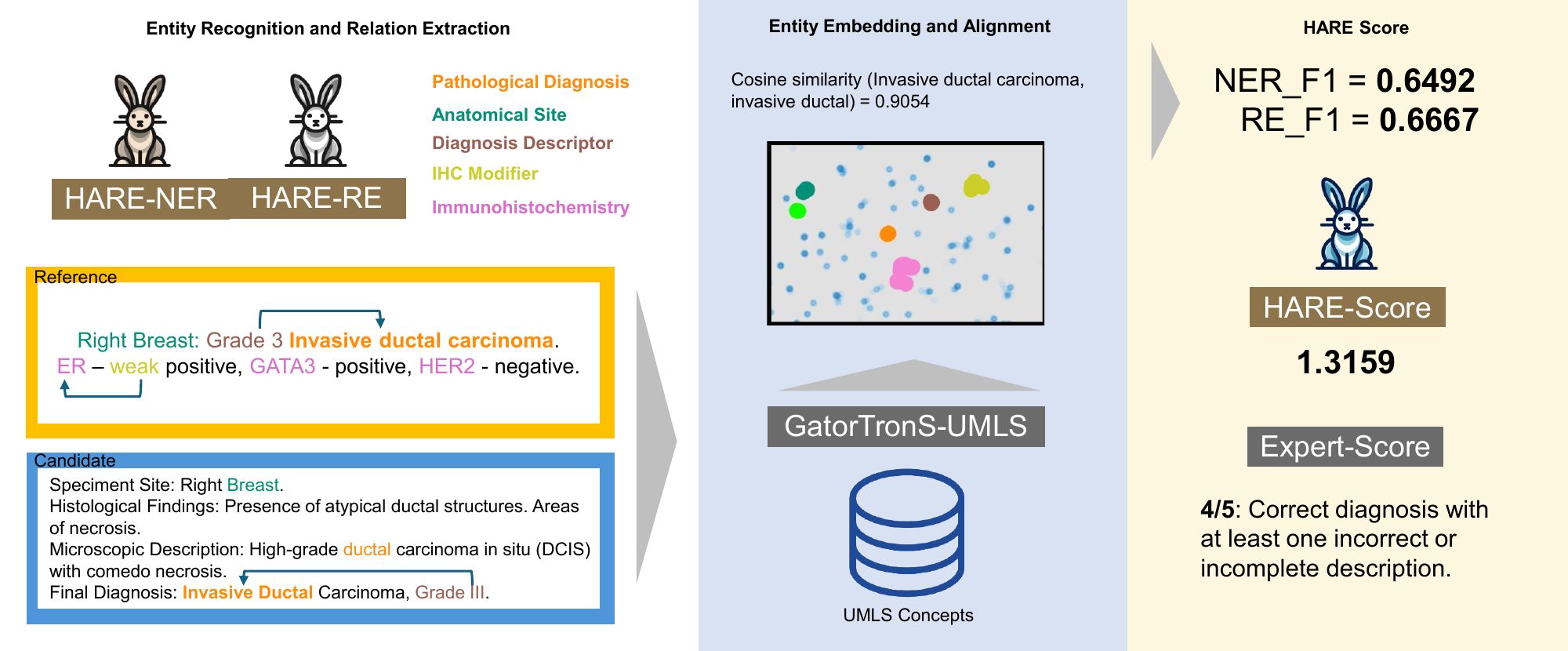}}
    \caption{Illustration of the process of computing the HARE score, a novel entity and relation centric metric for evaluating histopathology report generation.}
    \label{fig:hare}
\end{figure*}

To address this gap, we introduce HARE (Histopathology Automated Report Evaluation): a novel, entity-focused metric designed to assess the clinical quality of generated histopathology reports. In Figure \ref{fig:hare}, the process of computing the HARE score is demonstrated. HARE captures domain-specific entities (e.g., anatomical sites, IHC markers, descriptor and final diagnosis) and relationships between the entities from both candidate and reference reports and quantifies their alignment via a cosine similarity measure \cite{rahutomo2012cosinesim}. Our approach is grounded in a comprehensive annotation effort on 1,465 real-world diagnostic histopathology reports sourced from a large teaching hospital and from The Cancer Genome Atlas (TCGA) \cite{tomczak2015tcga}.

By emphasizing the presence and correctness of domain-specific entities, HARE provides a more clinically oriented benchmark than existing lexical metrics. We validated its effectiveness by demonstrating the higher correlation between HARE scores and expert-derived evaluations of generated reports compared with multiple other available metrics. By releasing both our annotated dataset and the final trained models (which we call HARE-NER and HARE-RE), we aim to encourage further research in histopathology NLP and to improve the clinical utility and reliability of automated report-generation systems.

The primary contributions of this paper are as follows:
\begin{enumerate}[noitemsep, topsep=0pt]
    \item \textbf{Introduction of a New Metric (HARE):} We propose a domain-specific evaluation metric for histopathology report generation that focuses on the detection and alignment of significant histopathology entities. To our knowledge, it is the first dedicated metric for this purpose.
    \item \textbf{Histopathology Score Dataset:} We collect and provide expert histopathologist scores for automatically-generated reports, demonstrating the real-world validity of HARE metric.
    \item \textbf{HARE-NER and HARE-RE:} We develop a NER model and a RE model specialized in histopathology, capable of recognizing and relating critical domain-specific entities such as IHC markers, anatomical sites and descriptor (for final diagnosis), filling a gap where there is no publicly available histopathology-focused NER model and RE model.
    \item \textbf{Open Source:} We will release (1) the annotated dataset, (2) the final trained NER model, RE model, as well as the alignment model, and (3) HARE score computation code to facilitate further research and development in both NER and report generation in the histopathology domain.
\end{enumerate}

\section{Related Work}
\label{sec:related_work}

While several evaluation metrics have been proposed for radiology, the field of histopathology remains underexplored. Two most recent notable contributions in radiology emphasize the design of domain-specific metrics that capture clinical significance: \textbf{RadGraph-XL} and \textbf{RaTEScore} \cite{jain2021radgraph,zhao2024ratescore}.

\subsection{RadGraph-XL}
RadGraph-XL \cite{delbrouck2024radgraphxl} is a large-scale, expert-annotated dataset created for extracting clinical entities and relations from radiology reports. Building upon its predecessor, RadGraph-1.0 \cite{jain2021radgraph}, RadGraph-XL expands annotations to cover multiple anatomy-modality pairs, including chest CT, abdomen/pelvis CT, and brain MRI, in addition to existing chest X-ray data. The dataset consists of over 2,300 reports annotated with 410,000 entities and relations, significantly enhancing its scale and diversity.

RadGraph-XL underscores the importance of clinically relevant entities and relationships in domain-specific metrics. This principle directly informs our work, as we extend it to the histopathology domain by focusing on uniquely critical entities such as features of the histopathological report including pathological diagnosis and IHC marker data.

\subsection{RaTEScore}
RaTEScore \cite{zhao2024ratescore} is a domain-specific evaluation metric designed to assess the quality of radiology report generation. Unlike general-purpose metrics such as BLEU or ROUGE, RaTEScore prioritizes clinical accuracy through entity-level assessments. It employs a NER module to extract key medical entities (e.g., anatomy, abnormalities, diseases) and a synonym disambiguation encoding module to address challenges such as medical synonymy and negation. The final metric is computed using the cosine similarity of entity embeddings, with adjustments made to reflect the clinical relevance of specific entity types.

To support its development, RaTEScore introduced two foundational resources:
\begin{enumerate}[noitemsep, topsep=0pt]
    \item \textbf{RaTE-NER}: A large-scale dataset for medical NER, covering nine imaging modalities and 22 anatomical regions.
    \item \textbf{RaTE-Eval}: A benchmark for evaluating metrics, including sentence- and paragraph-level human ratings, as well as comparisons involving synthetic reports.
\end{enumerate}

RaTEScore demonstrated superior alignment with human preferences, achieving the highest correlation scores in evaluations on public datasets such as ReXVal and the RaTE-Eval benchmark. Inspired by RaTEScore's methodology, our proposed HARE metric adapts the principles of entity recognition and embedding similarity to the histopathology domain, addressing unique challenges such as the interpretation of pathological diagnosis and IHC findings.

\subsection{Limitations in Existing Metrics}
Although RadGraph-XL and RaTEScore have significantly advanced the evaluation of radiology reports, their applicability is limited to specific modalities (e.g., chest X-rays) and radiological contexts. They do not address the unique linguistic and clinical knowledge of histopathology, which involve detailed morphological assessments and IHC findings. 

\textbf{HARE} addresses these limitations by introducing an entity-aware evaluation framework tailored specifically to the histopathology domain. By emphasizing the detection and alignment of domain-specific entities, HARE provides a clinically relevant metric to assess the quality of generated histopathology reports.

\section{Methods}
\label{sec:methods}

In this section, we describe the development of HARE (Histopathology Automated Report Evaluation), a domain-specific evaluation metric designed to assess the clinical quality of generated histopathology reports. Our methodology involves dataset preparation and annotation, NER model and RE model training, and the design of the HARE metric.

\subsection{Dataset Preparation and Annotation}
We curated two datasets to support the development of HARE: reports collected from a hospital and publicly available reports from TCGA.

\subsubsection{Hospital Dataset}
We collected 813 fully de-identified/anonymized histopathology reports from the pathology department of a large teaching hospital. We ensured that the reports were free of any identifiable data through the use of Stanford AIMI's deidentification model and by manual review and redaction of identifiers by three histopathologists \cite{stanford_deid}. The reports were from cases across a range of tissue types and diagnoses with a partial focus on cases with lymphoma, breast cancer and cases in which several IHC markers were utilized as part of the diagnostic process. The reports were annotated by a junior histopathologist (with input from a senior histopathologist for clarification of challenging cases) using the Inception annotation tool \cite{tubiblio106270_inception}. The annotations focused on histopathology-specific entities, including:
\begin{itemize}[noitemsep, topsep=0pt]
    \item \textbf{Anatomical Site:} Entities describing specific tissue regions or locations, such as \textit{breast, lung, kidney, lymph node etc.}
    \item \textbf{Immunohistochemistry (IHC) Markers:} The presence of immunohistochemical markers such as \textit{CK20, CDX2, ER, PR, Ki-67}.
    \item \textbf{Pathological diagnosis:} The pathological diagnosis, such as \textit{classical Hodgkin lymphoma}.
    \item \textbf{Diagnosis Descriptor:} Provides descriptive characteristics of the pathological diagnosis e.g., ``raises the possibility of''.
    \item \textbf{IHC Modifier:} Used to modify immunohistochemical annotations, e.g., ``patchy'' or ``strong''.
\end{itemize}
The relationships annotated were:
\begin{itemize}[noitemsep, topsep=0pt]
    \item \textbf{IHC Markers - IHC Modifier}
    \item \textbf{Diagnosis - Diagnosis Descriptor}
\end{itemize}

\begin{table}[htbp]
\centering
\begin{tabular}{lcc}
\hline
\textbf{Type} & \textbf{Hospital} & \textbf{TCGA} \\
\hline
IHC Markers & 6,628 & 119 \\
IHC Modifier & 1,339 & 173 \\
Pathological Diagnosis & 885 & 882 \\
Anatomical Site & 747 & 794 \\
Diagnosis Descriptor & 247 & 475 \\\hline
Relations & 1,745 & 653 \\\hline
\end{tabular}
\caption{Entity and Relation annotation statistics for the Hospital and TCGA datasets.}
\label{tab:entity_annotations}
\end{table}

\subsubsection{TCGA Dataset}
To increase diversity, we further annotated 652 publicly available histopathology reports from the previously published HistGen training and evaluation dataset, which is originally sourced from The Cancer Genome Atlas (TCGA) \cite{guo2024histgen,tomczak2015tcga}. The annotation was done in the same manner as the Hospital dataset but using the label studio as the annotation tool \cite{Label-Studio}. We extracted sentences with histopathological descriptions, specifically IHC markers and final diagnosis characteristics. The breakdown of the number of annotations for the Hospital and TCGA datasets are summarized in table \ref{tab:entity_annotations}.

\subsubsection{Annotator description}
All annotations were performed by practicing physicians with formal training and accreditation in histopathology (at different stages of progression through the national pathology examination board). The two junior pathologists had 5 and 7 years of histopathology experience, alongside 13 and 12 years of clinical medical practice, respectively. The senior pathologist was board-certified, with 7 years of histopathology experience and 18 years of medical practice. Before initiating the annotation process, all annotators met to establish and agree on a standardized annotation protocol. 

\subsection{HARE-NER and HARE-RE Training}
\begin{table}[htbp]
\centering
\begin{tabular}{p{5.0cm}p{1.85cm}}
\hline
\textbf{General Domain} & \textbf{Model Size}\\
\hline
BERT\cite{devlin2018bert} & 110M 340M \\
DeBERTa-v3\cite{he2021debertav3} & 70M 435M \\
\hline
\textbf{Biomedical Domain} & \textbf{Model Size}\\
\hline
PathologyBERT\cite{santos2023pathologybert} & 110M \\
BiomedBERT\cite{BioMedBERTandELECTRA} & 110M 340M \\
SapBERT\cite{liu2020sapbert} & 110M \\
GatorTronS\cite{yang2022gatortron} & 345M \\ \hline
\end{tabular}
\caption{Models tested for fine-tuning. The models are sorted in the order of size. Models with two sizes indicate different pretrained model variants (e.g., BERT-base vs. BERT-large).}
\label{tab:models}
\end{table}

As shown in Table \ref{tab:models}, we experimented with several transformer-based architectures, including PathologyBERT \cite{santos2023pathologybert} and GatorTronS \cite{yang2022gatortron}, which are pre-trained on clinical corpora, and BiomedBERT \cite{BioMedBERTandELECTRA} which was trained with PubMed articles as well as general domain models (BERT \cite{devlin2018bert} and DeBERTa \cite{he2021debertav3} models). PathologyBERT is the only publicly available model that is trained with pathology reports for document classification specifically for breast cancer. SapBERT \cite{liu2020sapbert} is also included as it was further trained with BiomedBERT model for entity alignment to Unified Medical Language System (UMLS), a detailed and widely used ontology \cite{UMLS_2024}. 

These models were fine-tuned using our annotated dataset for both NER and RE. For the NER task, we trained a token classification model based on the pre-trained encoder to recognize histopathology-specific entities. For the RE task, we trained a sequence classification model with entity markers (E1 and E2) based on the same pre-trained encoders to capture relationships between extracted entities. E1 and E2 are placeholder tokens used to mark the two extracted entities involved in a candidate relation pair (they are abbreviations for entity 1 and entity 2, respectively). These are passed to the relation extraction model, which classifies the relation type via a sequence classification objective.

The annotated reports were split into sentences, and any sentence longer than 512 tokens was split during preprocessing to accommodate model input constraints. All models were implemented using the HuggingFace Transformers library \cite{wolf2019huggingface}. Training was conducted on an NVIDIA A5000 GPU. For both NER and RE, we used an AdamW optimizer with a learning rate of $5\times10^{-5}$ and a batch size of 4 for 2 epochs. Evaluation was performed using standard metrics, F1-score, for both tasks, with 10\% of the data as a hold-out test set. 

For relation extraction, the dataset required explicit construction of entity pairs. All positive samples (annotated entity relations) and an equal number of randomly sampled negative pairs were used to construct the training split. For the test split, three times as many negative samples as positive samples were sampled to ensure robust evaluation. The relation extraction model was evaluated using gold-standard entities, not predicted ones. 

Details of the train and test splits are shown in Table~\ref{tab:combined_dataset_stats}. The best-performing models for NER and RE were selected as the backbone for extracting entities and relationships within the HARE metric.

\begin{table}[htbp]
\centering
\begin{tabular}{lcccc}
\hline
\textbf{Split} & \textbf{Samples} & \textbf{Tokens} \\
\hline
NER-Train & 2,181 & 127,553 \\
NER-Test & 243 & 13,855 \\\hline
Relation-Train & 5,014 & 311,058 \\
Relation-Test & 1,068 & 66,769 \\\hline
\end{tabular}
\caption{Statistics of the train and test datasets used for NER and RE tasks. \textbf{Samples} represents the number of samples and \textbf{Tokens} the total tokens (word-piece).}
\label{tab:combined_dataset_stats}
\end{table}

\subsection{Design of the HARE Metric}
The HARE metric evaluates the quality of generated histopathology reports by assessing both the alignment of clinically relevant entities and the relationships between them in the reference and candidate reports.

\subsubsection{Entity and Relation Extraction}
Entities are extracted from both reference and candidate reports using the trained HARE-NER model. For each token, the model outputs a probability distribution over entity classes; only entities with confidence scores above a threshold of 0.7 are retained, ensuring that uncertain predictions are excluded. Relations between recognized entities are then identified using the trained HARE-RE model, which predicts relation types for all candidate entity pairs. The same confidence threshold is applied to relation predictions to retain only high-confidence relations.

\subsubsection{Entity Embedding and Alignment}
Extracted entities are embedded in a high-dimensional space using contextual representations from GatorTronS, further fine-tuned with a UMLS-based SapBERT approach to ensure semantic alignment of similar entities (e.g., \textit{lymphovascular invasion} and \textit{vascular invasion}). Cosine similarity is computed between all entity pairs from reference and candidate reports. For each entity, the maximum cosine similarity with entities in the other set is calculated.

\subsubsection{Scoring}
The HARE metric reports both entity- and relation-level alignment between candidate and reference reports. For entities, precision, recall, and F1-score are computed as follows:

\vspace{-1.5em}
\begin{align*}
\text{Recall}_{\text{e}} = \frac{1}{|\mathbf{E}_{\text{ref}}|} \sum_{\mathbf{e}_{\text{ref}} \in \mathbf{E}_{\text{ref}}} \max_{\mathbf{e}_{\text{cand}} \in \mathbf{E}_{\text{cand}}} \text{S}(\mathbf{e}_{\text{ref}}, \mathbf{e}_{\text{cand}})
\end{align*}
\vspace{-1.5em}
\begin{align*}
\text{Precision}_{\text{e}} = \frac{1}{|\mathbf{E}_{\text{cand}}|} \sum_{\mathbf{e}_{\text{cand}} \in \mathbf{E}_{\text{cand}}} \max_{\mathbf{e}_{\text{ref}} \in \mathbf{E}_{\text{ref}}} \text{S}(\mathbf{e}_{\text{cand}}, \mathbf{e}_{\text{ref}})
\end{align*}
where $\mathbf{E}_{\text{ref}}$ and $\mathbf{E}_{\text{cand}}$ are the sets of embeddings for reference and candidate entities, and $\text{S}(\mathbf{u}, \mathbf{v})$ is the cosine similarity between embeddings $\mathbf{u}$ and $\mathbf{v}$.

The F1-score for NER is then calculated as the harmonic mean of precision and recall:

\vspace{-1.5em}
\begin{align*}
\text{F1}_{\text{e}} = 2 \cdot \frac{\text{Precision}_{\text{e}} \cdot \text{Recall}_{\text{e}}}{\text{Precision}_{\text{e}} + \text{Recall}_{\text{e}}}
\end{align*}
\vspace{-1em}

Relation extraction performance is quantified using the standard F1-score, computed by comparing the set of extracted relations (entity pairs and their predicted relation types) in the candidate report to those in the reference. Precision and recall are calculated based on the overlap of predicted and reference relations, and the relation F1-score is reported as:

\vspace{-1.5em}
\begin{align*}
\text{F1}_{\text{r}} = 2 \cdot \frac{\text{Precision}_{\text{r}} \cdot \text{Recall}_{\text{r}}}{\text{Precision}_{\text{r}} + \text{Recall}_{\text{r}}}
\end{align*}
\vspace{-1em}

To obtain a comprehensive assessment, the final HARE score is defined as the sum of the entity and relation F1-scores:

\vspace{-1.5em}
\begin{align*}
\text{HARE Score} = \text{F1}_{\text{e}} +  \text{F1}_{\text{r}}
\end{align*}
\vspace{-1.5em}

This ensures that both precision and recall are considered equally, providing a balanced measure of the alignment between ground truth and predicted entities. A higher HARE score indicates better alignment, reflecting both accurate and comprehensive entity matching.

\subsection{Validation of the HARE Metric}
To validate HARE, we conducted an expert evaluation of machine-generated histopathology reports. We generated reports using GPT-4o and GPT-4o-mini using whole slide images (WSI) downloaded from TCGA \cite{hurst2024gpt}. Due to the volume of the images, we processed to lower resolution and resized the image to 1024 by 1024 pixels. In total, 75 randomly selected images were downloaded and used for generating reports. For each image, eight sets of reports were generated with different levels of specimen site information provided. In total, 600 reports were compared to the ground truth reports. Experts compared generated reports to ground truth (original) reports and assigned scores based on diagnostic accuracy and histopathological detail to ensure an objective evaluation of the model's performance in generating histopathology reports from WSI. 

The following is the scoring system and the rationale for each score level: 

\begin{itemize}[noitemsep, topsep=0pt]
    \item \textbf{Scores 5 (Perfect match with ground truth)}: This score is assigned to reports that are identical to the reference report in terms of both diagnostic accuracy and histopathological descriptions. 
    \item \textbf{Scores 4 (Perfect match diagnosis with at least one wrong description)}: This score is assigned to reports that correctly identify the diagnosis, but contain at least one minor error in histopathological descriptions. These errors may involve inaccurate terminology or missing morphological features. Although these reports provide a reliable diagnosis, an incomplete or incorrect description reduces their overall quality. 
    \item \textbf{Scores 3 (Correct diagnosis)}: This score is assigned to reports that accurately determine the correct diagnosis but do not provide any of the detailed histopathological descriptions in the ground truth. 
    \item \textbf{Scores 2 (Broadly correct diagnosis)}: This score is assigned when reports correctly identify the general disease category but do not specify the exact diagnosis. For example, a report may correctly classify a tumor as malignant but does not differentiate between specific subtypes. These reports provide a useful but incomplete diagnosis, which limits their clinical applicability. 
    \item \textbf{Scores 1 (Incorrect diagnosis with some of the histopathological descriptions matching the ground truth)}: This score is assigned when the report fails to provide the correct diagnosis but includes histopathological descriptions that align with the reference report. While some microscopic features are correctly described, the overall diagnostic conclusion is incorrect, greatly reducing the clinical reliability and utility of the report. 
    \item \textbf{Scores 0 (Incorrect diagnosis with no histopathological descriptions matching with ground truth )}: This score is assigned to reports that provide neither a correct diagnosis nor any histopathological descriptions that align with the ground truth. These reports fail to recognize key pathological features and do not contribute to an accurate clinical assessment, making them completely unreliable. 
\end{itemize}

Histopathology reports are inherently complex and exhibit significant variability in writing styles across institutions and individual histopathologists, particularly in the microscopic description section. This variability introduces heterogeneity in report structure, making it challenging for models to learn consistent diagnostic patterns. Despite these differences, histopathologists generally reach a consensus on the final diagnosis, which carries the most clinical significance. Therefore, our evaluation places greater emphasis on the model’s ability to generate correct diagnoses rather than the accuracy of microscopic descriptions.

\subsubsection{Metric Comparison}
HARE scores were compared to expert scores using Pearson’s correlation coefficient, Spearman's correlation coefficient, and Kendall's $\tau$. We provide p-values for each correlation value. Additionally, we benchmarked the metric against traditional lexical metrics (BLEU, ROUGE, METEOR, BERTScore) and radiology-specific metrics (RadGraph-XL, RaTEScore, GREEN) \cite{papineni2002bleu,lin2004rouge,banerjee2005meteor,zhang2019bertscore,delbrouck2024radgraphxl,zhao2024ratescore,ostmeier-etal-2024-green}. We used only the single overall score for metrics such as BLEU, RadGraph-XL, and GREEN to compare our method with a state-of-the-art LLM-based evaluator for radiology reports. We also used GPT-4.1 as a judge to give a score of 0-5 based on the expert evaluation scheme for the candidate report based on the ground truth report \cite{hurst2024gpt}. The prompt used for this analysis can be found at the appendix Figure \ref{fig:gpt4.1prompt}. We also performed regression analysis and provided R$^2$ and RMSE values to assess the predictive utility of each metric against expert scores. For all the metric comparison, we normalized the automated metric scores to a 0-1 scale and expert evaluation scores (originally 0–5) also normalized to 0-1. 

\section{Results and Discussion}
\label{sec:results_discussion}

\subsection{Model Selection: GatorTronS}

\begin{table}[tbhp]
\centering
\begin{tabular}{lccc}
\hline
\textbf{Model} & \textbf{NER} & \textbf{RE} & \textbf{Overall} \\
\hline
PathologyBERT & 0.771 & 0.798 & 0.785 \\
BERT-base & 0.833 & 0.798 & 0.816 \\
BERT-large & 0.825 & 0.798 & 0.811 \\
DeBERTa-large & 0.841 & 0.798 & 0.820 \\
BiomedBERT-large & 0.843 & 0.798 & 0.820 \\ 
DeBERTa-xsmall & 0.794 & 0.962 & 0.878 \\
SapBERT & 0.835 & 0.970 & 0.903 \\
BiomedBERT-base & 0.844 & 0.962 & 0.903 \\ \hline
\textbf{GatorTronS} & \textbf{0.854} & \textbf{0.977} & \textbf{0.915}\\\hline
\end{tabular}
\caption{Model selection results based on NER and RE F1-scores on the test set. Models are sorted by Overall F1-score.}
\label{tab:model_selection}
\end{table}

Our experiments demonstrated that GatorTronS outperforms other models, both general-purpose and biomedical, in extracting entities and relations from histopathology reports. As shown in Table~\ref{tab:model_selection}, GatorTronS achieved the highest overall score (0.915) with NER F1 (0.854) and RE F1-score (0.977), surpassing the next-best model, BiomedBERT-base (Overall F1 = 0.903, NER F1 = 0.844, RE F1 = 0.962). Notably, models with an RE F1 of 0.798 failed to identify any relations for all test inputs, highlighting a limitation of these architectures in this domain. 

This result underscores the efficacy of GatorTronS in addressing the complexities inherent in histopathology text. Its extensive pre-training on large-scale synthetic clinical corpora provides it with a comprehensive understanding of domain-specific language, abbreviations, and nuanced terminology. This ability is particularly critical in histopathology, where specialized expressions describing tissue morphology and disease subtypes are prevalent. 

An additional factor contributing to GatorTronS's superior performance is its model size. As the largest model among the biomedical models tested, GatorTronS benefits from greater representational capacity, enabling it to capture complex relationships in text more effectively.

\subsection{Majority of Generated Reports Lack Clinical Alignment}

\begin{table}[t]
\centering
\begin{tabular}{lc}
\hline
\textbf{Score} & \textbf{Count} \\
\hline
0 & 369 \\
1 & 71 \\
2 & 90 \\
3 & 62 \\
4 & 8 \\
5 & 0 \\
\hline
\end{tabular}
\caption{Distribution of expert evaluation scores for generated histopathology reports. Scores represent the degree of alignment with the reference reports, with higher scores indicating better alignment.}
\label{tab:expert_evaluation}
\end{table}

Despite advances in text generation methods, expert evaluations reveal a significant misalignment between system-generated reports and clinical requirements. As shown in Table~\ref{tab:expert_evaluation}, 369 out of 600 generated reports (61.5\%) received a score of 0 and 71 reports received a score of 1 (11.8\%), indicating 73.3\% of the reports had an incorrect diagnosis. Only eight reports attained a score of 4, while none achieved the perfect score of 5. Scores with partially correct diagnosis, broadly correct diagnosis, and correct diagnosis (Score 2, 3, and 4) accounted for 160 reports (26.7\%). When we compared the HARE and other scores to expert scores, we excluded reports with 0 scores to have more balanced representation of the scores. Reports often lacked diagnostic conclusions or included incorrect terminology, while others failed to capture essential histological findings. The scarcity of high-quality outputs underscores the challenge in generating nuanced and diagnostically accurate narratives. Errors in final diagnosis are particularly concerning as they can have significant clinical implications. These findings highlight a significant limitation in the diagnostic accuracy of the generative model utilized, with a substantial proportion of reports failing to predict reliable pathological interpretations.

The high percentage of incorrect diagnoses and the lack of accurate microscopic descriptions can be attributed to several factors. One major limitation can be the use of a single, low-resolution WSI, which could restrict the model's ability to discern detailed morphological features essential for histopathological evaluation. Histopathologists analyze WSIs at multiple magnification levels (low-power magnification for architectural patterns, high-power for cellular details such as nuclear atypia, and mitotic figures), which is crucial to make an accurate pathological diagnosis. This limitation can hinder the model's capacity to generate precise microscopic descriptions and accurately differentiate pathological entities. Furthermore, only one WSI was provided per case, whilst in most cases multiple WSIs were utilized as part of the actual diagnostic process to generate the ground truth report. Finally, critical contextual information (e.g., clinical history or anatomical site information) was not provided all the time. Notably, a subset of reports that included primary specimen site information demonstrated a slight improvement, achieving higher scores. This suggests that while the performance of current multimodal LLMs such as GPT-4o, is limited, when provided with additional clinical and anatomical context, the model's diagnostic reliability can sometimes be acceptable.

\begin{table*}[tbhp]
\centering
\resizebox{\textwidth}{!}{
\begin{tabular}{lcccccccc}
\hline
\textbf{Method} & \textbf{$r$} & \textbf{$r$ p-val} & \textbf{$\rho$} & \textbf{$\rho$ p-val} & \textbf{$\tau$} & \textbf{$\tau$ p-val} & \textbf{R$^2$} & \textbf{RMSE} \\
\hline
ROUGE-L      & 0.048   & 0.470     & 0.030   & 0.647     & 0.025   & 0.616     & 0.002   & 0.169 \\
BLEU         & 0.077   & 0.241     & 0.106   & 0.108     & 0.099   & 0.107     & 0.006   & 0.168 \\
GPT-4.1	& 0.177	& 0.007	& 0.173	& 0.008 &	0.146 &	0.008	& 0.031 &	0.166 \\
BERTScore    & 0.203   & 0.002     & 0.180   & 0.006     & 0.141   & 0.005     & 0.041   & 0.165 \\
METEOR       & 0.265   & 4.51e-05  & 0.179   & 0.006     & 0.136   & 0.007     & 0.070   & 0.163 \\
RaTEScore    & 0.372   & 5.36e-09  & 0.350   & 4.81e-08  & 0.276   & 4.60e-08  & 0.138   & 0.157 \\
RadGraph-XL  & 0.427   & 1.22e-11  & 0.425   & 1.43e-11  & 0.351   & 8.51e-11  & 0.182   & 0.153 \\
GREEN        & 0.438   & 2.90e-12  & 0.482   & 7.58e-15  & 0.410   & 2.18e-13  & 0.192   & 0.152 \\
\textbf{HARE (Ours)}  & \textbf{0.606} & \textbf{1.39e-24} & \textbf{0.643} & \textbf{2.62e-28} & \textbf{0.533} & \textbf{1.51e-24} & \textbf{0.368} & \textbf{0.134} \\
\hline
\end{tabular}
}
\caption{Comparison of evaluation methods based on Pearson correlation ($r$), Spearman ($\rho$), and Kendall's $\tau$ with their $p$-values, and regression performance (R$^2$ and RMSE). Methods are sorted by Pearson correlation $r$.}
\label{tab:evaluation_comparison}
\end{table*}

\subsection{HARE Outperforms Existing Metrics in Capturing Clinical Relevance}

Table~\ref{tab:evaluation_comparison} summarizes the performance of all evaluation metrics against expert pathologist scores using multiple statistical measures. HARE achieved the highest Pearson correlation (0.606), Spearman correlation (0.643) and Kendall $\tau$ (0.533), all with strong statistical significance. HARE also demonstrated the highest coefficient of determination ($R^2 = 0.368$) and the lowest root mean squared error (RMSE = 0.134), indicating both high alignment and predictive accuracy with respect to expert score. 

These results surpass those of GREEN, the next-best metric, which leverages a large language model (RadLlama2-7b) as an evaluator. Moreover, HARE is significantly more computationally efficient: on 600 candidate reports, GREEN required 2 hours and 2 minutes for evaluation, while HARE completed the same analysis in 192 seconds on an A5000 24GB GPU. This efficiency, combined with robust performance, underscores HARE’s practical viability and interpretability as an evaluation metric for histopathology report generation.

HARE significantly outperforms GPT-4.1 used as a judge in correlation with expert ratings. While GPT-4.1 is not appropriate for real clinical evaluation pipelines due to its proprietary nature and privacy constraints, our result confirms the superior performance of a domain specific report quality evaluator over an LLM based evaluator.

In contrast, although they are widely used in histopathology report evaluation, lexical metrics such as ROUGE-L ($r = 0.048$, $\rho = 0.030$, $\tau = 0.025$) and BLEU ($r = 0.078$, $\rho = 0.106$, $\tau = 0.099$) showed minimal correlation and high RMSE, further underscoring their inability to assess clinically relevant content in histopathology. 

HARE’s effectiveness originates from its focus on histopathology entity-level alignment, which ensures that key clinical features, such as pathological diagnosis, are appropriately prioritized. Unlike traditional lexical metrics, HARE incorporates semantic similarity measures tailored to pathology-specific terminology by incorporating descriptor and modifier entities, making it robust to linguistic variations. By capturing both semantic and clinical correctness, HARE offers a more accurate and reliable evaluation of generated histopathology reports.

The implications of HARE’s performance are significant. Its strong correlation with expert evaluations indicates that it is a reliable proxy for clinical relevance and accuracy of the generated reports. HARE can guide iterative improvements in report generation models, ensuring that future systems better align with clinical requirements.

\section{Conclusion} \label{sec:conclusion}

In this work, we proposed HARE, a novel entity and relation centric evaluation metric specifically designed to assess the clinical quality of machine-generated histopathology reports. HARE addresses the critical gap in domain-specific evaluation by prioritizing clinical relevance. HARE effectively aligns with expert evaluations, outperforming existing metrics such as ROUGE and RaTEScore.

Our findings reveal that even proprietary multimodal large language models, such as GPT-4o, struggle to produce clinically accurate histopathology reports. Although we have not tested a comprehensive list of models trained for histopathology reports such as HistGen and WsiCaption, HARE can be a robust framework for evaluating these models \cite{guo2024histgen,chen2024wsicaption}. In the future, we will include histopathology reports specific models for creating the human evaluation dataset as well as extend HARE models to a joint NER+RE model to further improve the performance and utility.

HARE's superior performance underscores the importance of domain-specific evaluation metrics in bridging the gap between automated report generation and clinical expectations. By making HARE publicly available, along with the annotations and models, we aim to facilitate advancements in both report generation and evaluation methodologies in histopathology and related fields.

\section*{Limitation} \label{sec:limitation}

\textbf{Scope of clinical entities and relations:}
The current implementation of HARE primarily addresses a set of core histopathology entities and relatively simple binary relations. More nuanced or higher-order clinical relationships, as well as rare or emerging entity types, remain underrepresented. Expanding both the entity and relation taxonomies to better reflect the complexity of real-world histopathology reporting is an important and interesting future work we plan to explore.

\textbf{Negation and uncertainty handling:}
While HARE captures explicit clinical entities, it does not yet explicitly handle negation or uncertainty (e.g., ``no evidence of malignancy,'' ``cannot rule out invasion''). These linguistic phenomena are important for accurate clinical interpretation and could be incorporated into future extensions of the metric.

\textbf{Breadth of expert evaluation models:}
For the generation of reports used in the expert evaluation, we utilized only two closed source models, GPT-4o and GPT-4o-mini. As the primary scope of this work was the development of the evaluation metric, a broader evaluation across more generative models remains to be explored in future work. 

\section*{Broader Impacts and Ethics Statement} \label{sec:ethics}

Histopathology reports used in this work were provided via a study registered with, and approved by, the NHS Health Research Authority (references 293404 and 23/LO/0253). All histopathology reports were fully de-identified to protect patient privacy and ensure compliance with ethical standards. No personally identifiable information was used in the development of the HARE framework. Our work does not raise any major ethical concerns. HARE is designed for evaluation and research purposes only and is not intended for direct use in clinical decision-making.

While HARE provides a reliable metric for evaluating the quality of generated histopathology reports, it does not address potential biases or hallucinations in the underlying text generation models. Therefore, any use of automated text generation systems in clinical workflows should include rigorous human oversight to mitigate risks, such as incorrect diagnoses or misleading conclusions.

\section*{Acknowledgement}
The authors kindly acknowledge funding from an NIHR Academic Clinical Lectureship (APL), The Jean Shanks Foundation/The Pathological Society of Great Britain \& Ireland (APL and MWSO) and Rosetrees Trust (APL and YK).

The results shown here are in part based upon data generated by the TCGA Research Network: \url{https://www.cancer.gov/tcga}.

\clearpage
\bibliography{acl_latex}

\clearpage
\appendix
\section*{Appendix}
\label{sec:appendix}

\section{Word cloud representations of radiology and histopathology reports}

\begin{figure}[htbp]
    \centering
    \includegraphics[trim={0cm 0cm 0cm 0cm},clip,width=0.5\textwidth]{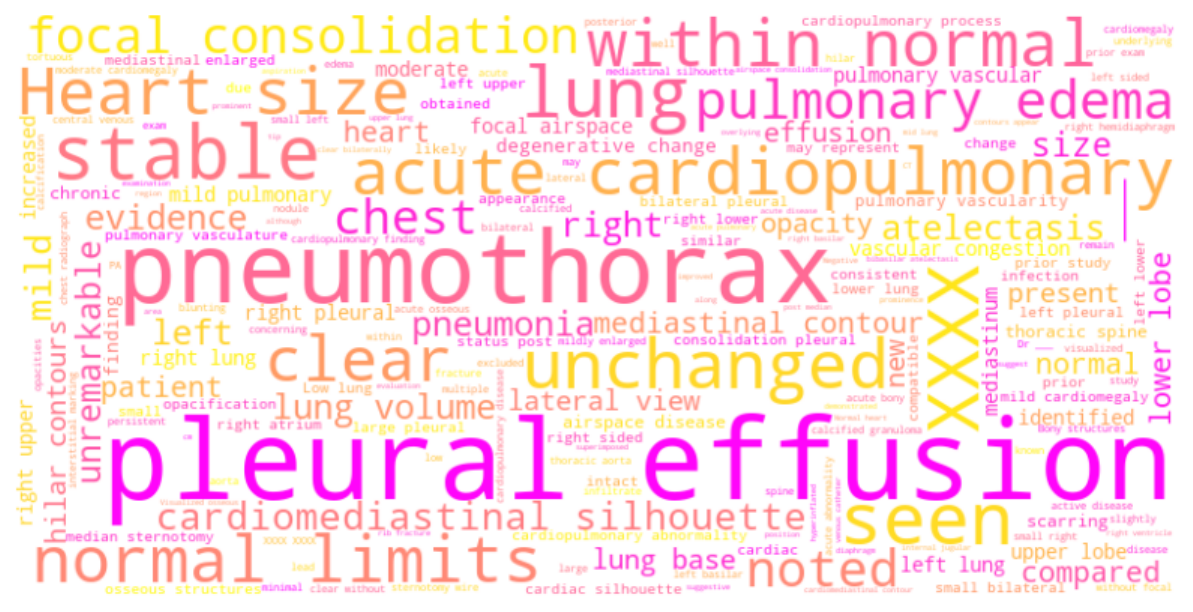}
    \caption{Word clouds of radiology reports. The radiology reports are 1,000 randomly sampled reports from MIMIC-CXR dataset and IU-X-ray dataset \cite{johnson2019mimic,Dina2016IU-Xray}. The size of each word represents its relative frequency in the corresponding category.}
    \label{fig:wc_radiology}
\end{figure}

\begin{figure}[htbp]
    \centering
    \includegraphics[trim={0cm 0cm 0cm 0cm},clip,width=0.5\textwidth]{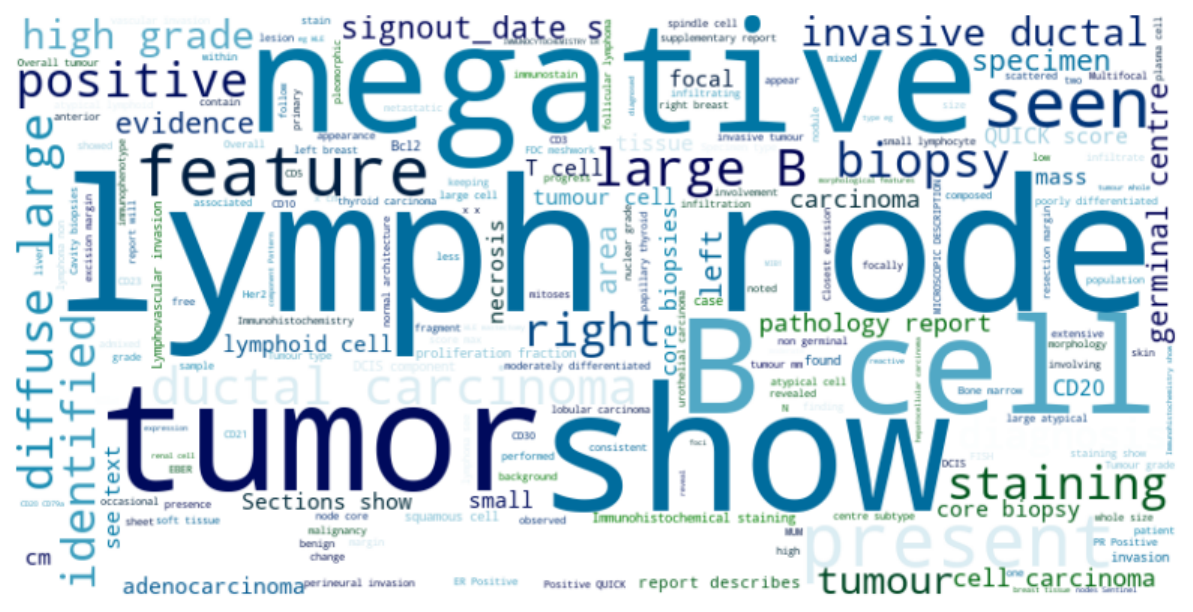}
    \caption{Word clouds of histopathology reports. The histopathology reports are 1,000 randomly sampled reports from our datasets. The size of each word represents its relative frequency in the corresponding category.}
    \label{fig:wc_histopathology}
\end{figure}

These visualizations provide insight into the linguistic differences between radiology and histopathology reports, highlighting the specialized vocabulary and diagnostic focus within each domain. Larger words represent higher relative frequency. 
The word cloud visualization for radiology reports highlights key terms such as "pleural effusion", "pneumothorax", "cardiopulmonary" and "atelectasis", indicating these are more common findings and diagnostic terminology used in radiology (see Figure \ref{fig:wc_radiology}). Figure \ref{fig:wc_histopathology} illustrates a word cloud generated from 1,000 randomly sampled histopathology reports from our datasets. Frequent occurring terms such as "tumor", "lymph node", "B cell", "negative", "biopsy", and "staining", reflect key features and diagnostic language used in histopathology reports. Compared to radiology reports, histopathology reports exhibit more granular terminology related to cellular morphology and pathology-specific descriptors.

\section{Report Examples}
We provide example annotated histopathology reports from both the Hospital and TCGA datasets. These examples illustrate not only the complexity and diversity of histopathology reporting, but also the breadth of clinically significant entities and inter-entity relationships captured by our annotation schema. Key entity types include pathological diagnosis, anatomical site, histological findings, immunohistochemistry markers, descriptors, and modifiers.

In addition to highlighting individual entities, these examples also depict the relationships between entities, such as associations between anatomical sites and diagnostic findings, or between immunohistochemistry results and corresponding pathological diagnoses. Modeling both entity-level information and their relationships is essential for accurately representing the clinical reasoning process in histopathology and for evaluating the fidelity of automated report generation.

Visualizing these examples demonstrates the level of annotation granularity and relational structure necessary for effective evaluation, and serves as a benchmark for downstream clinical NLP applications in entity recognition and relation extraction.

\begin{figure*}
    \centering
    \fbox{\includegraphics[width=1.0\linewidth]{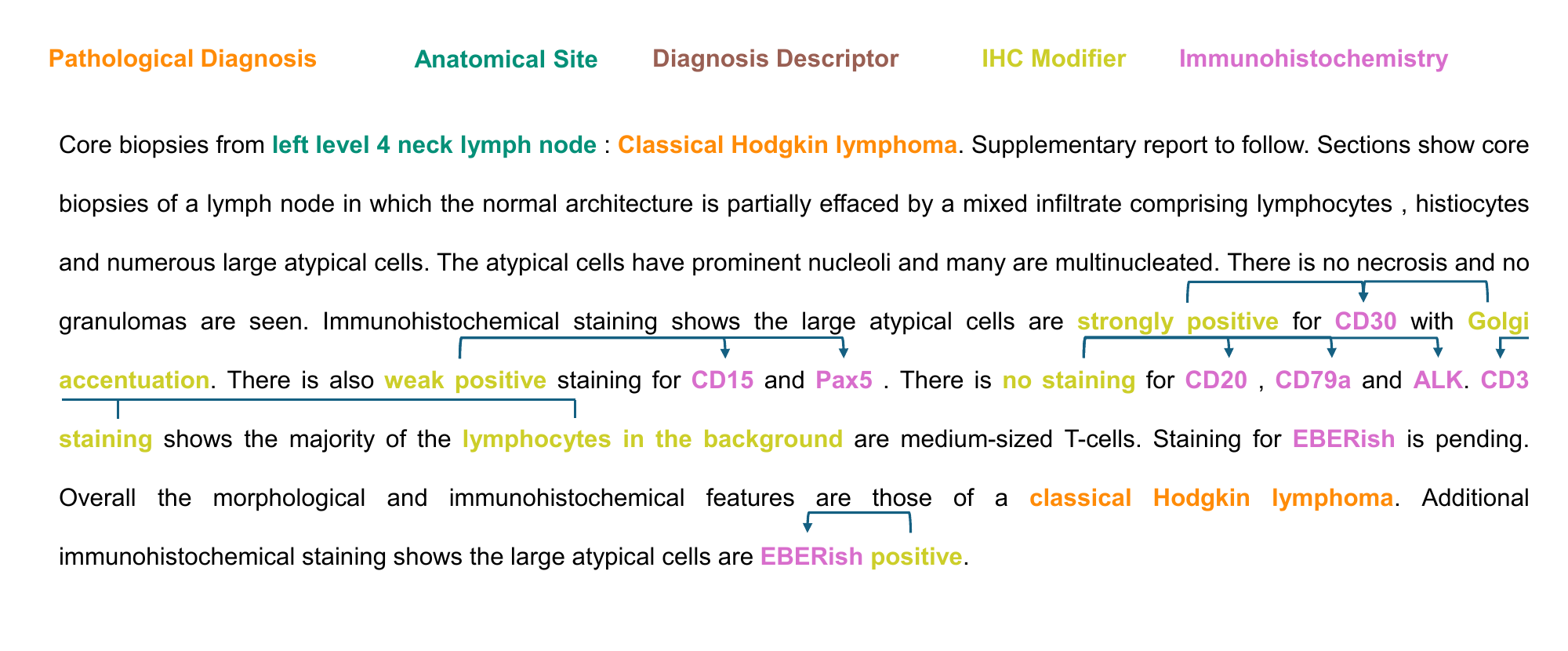}}
    \caption{Example of an annotated histopathology report from the Hospital Dataset. The report details a diagnosis of classical Hodgkin lymphoma in a lymph node, with corresponding entity-level annotations highlighting pathological diagnosis, anatomical site, immunohistochemical findings, and key descriptors.}
    \label{fig:hospital-example}
\end{figure*}
\begin{figure*}
    \centering
    \fbox{\includegraphics[width=1.0\linewidth]{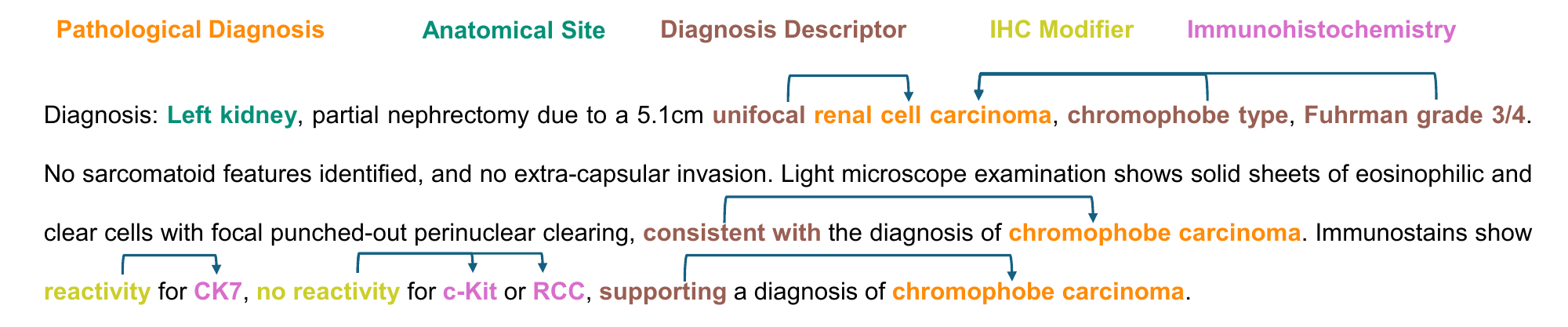}}
    \caption{Example of an annotated histopathology report from the TCGA Dataset. The report presents a case of left kidney partial nephrectomy for chromophobe renal cell carcinoma, with entity annotations for pathological diagnosis, anatomical site, diagnostic descriptors, immunohistochemistry markers, and modifiers.}
    \label{fig:tcga-example}
\end{figure*}

\begin{table*}[tbhp]
\centering
\resizebox{\textwidth}{!}{
\begin{tabular}{lcccccccc}
\hline
\textbf{Method} & \textbf{$r$} & \textbf{$r$ p-val} & \textbf{$\rho$} & \textbf{$\rho$ p-val} & \textbf{$\tau$} & \textbf{$\tau$ p-val} & \textbf{R$^2$} & \textbf{RMSE} \\
\hline
HARE\_ERROR & 0.026	& 0.693 &	-0.005 &	0.942	& -0.005 &	0.934 &	0.001 & 	0.169 \\
HARE\_No\_Threshold  & 0.567 & 4.89e-21 & 0.629 & 7.86e-27 & 0.513 & 4.00e-24 & 0.321 & 0.139 \\
\textbf{HARE\_0.7\_Threshold} & \textbf{0.606} & \textbf{1.39e-24} & \textbf{0.643} & \textbf{2.62e-28} & \textbf{0.533} & \textbf{1.51e-24} & \textbf{0.368} & \textbf{0.134} \\
\hline
\end{tabular}
}
\caption{Comparison of evaluation methods based on Pearson correlation ($r$), Spearman ($\rho$), and Kendall's $\tau$ with their $p$-values, and regression performance (R$^2$ and RMSE). Methods are sorted by Pearson correlation $r$. HARE\_ERROR is the one with inverted confidence threshold. HARE\_No\_Threshold is the one without threshold. HARE\_0.7\_Threshold is our method.}
\label{tab:evaluation_ablation}
\end{table*}

\section{Empirical anslysis of NER and RE errors}

To empirically assess the impact of errors in NER and RE, we conducted an ablation study evaluating several variants of the HARE metric under different entity and relation confidence thresholding schemes. Specifically, we compared:
\begin{enumerate}[noitemsep, topsep=0pt]
    \item \textbf{HARE\_ERROR}: HARE applied using only low-confidence (i.e., likely incorrect) NER and RE outputs by inverting the threshold
    \item  \textbf{HARE\_No\_Threshold}: HARE applied with no confidence threshold, including all predicted entities and relations
    \item \textbf{HARE\_0.7\_Threshold}: our default approach, which applies a confidence threshold of 0.7 to retain only high-confidence entities and relations
\end{enumerate}

Table ~\ref{tab:evaluation_ablation} presents the results. The HARE\_ERROR variant demonstrates very poor correlation with expert scores across all statistical measures, underscoring the critical importance of accurate entity recognition and relation extraction. Removing the threshold altogether (HARE\_No\_Threshold) moderately improves performance but still underperforms relative to our approach. The HARE\_0.7\_Threshold, our approach, achieves the highest correlation and lowest RMSE, validating our choice of threshold and the metric’s design, which effectively mitigates the impact of noisy or uncertain predictions.

These findings highlight that HARE’s strong correlation with expert assessments depends critically on accurate entity recognition and relation extraction. It also shows that the chosen confidence thresholding scheme is a key mechanism for maintaining robustness to NER and RE errors.

\section{GPT4.1 Prompt}
We designed the prompt for GPT4.1 analysis for the expert evaluation of the machine-generated histopathology reports \ref{fig:gpt4.1prompt}. 
\begin{figure*}[thbp!]
\centering
\begin{tcolorbox}[title= 
 Prompt for GPT-4.1, width=\textwidth, boxsep=1mm, left=1mm, right=1mm, top=1mm, bottom=1mm,]
Act as a histopathologist.\\
Review the following candidate report's similarity score to the ground truth report. Just report the numerical score.\\
The following is the scoring system and the rationale for each score level:\\
 - Scores 5 (Perfect match with ground truth): This score is assigned to reports that are identical to the reference report in terms of both diagnostic accuracy and histopathological descriptions. \\
 - Scores 4 (Perfect match diagnosis with at least one wrong description): This score is assigned to reports that correctly identify the diagnosis, but contain at least one minor error in histopathological descriptions. These errors may involve inaccurate terminology or missing morphological features. Although these reports provide a reliable diagnosis, an incomplete or incorrect description reduces their overall quality. \\
 - Scores 3 (Correct diagnosis): This score is assigned to reports that accurately determine the correct diagnosis but do not provide any of the detailed histopathological descriptions in the ground truth.\\ 
 - Scores 2 (Broadly correct diagnosis): This score is assigned when reports correctly identify the general disease category but do not specify the exact diagnosis. For example, a report may correctly classify a tumor as malignant but does not differentiate between specific subtypes. These reports provide a useful but incomplete diagnosis, which limits their clinical applicability.\\
 - Scores 1 (Incorrect diagnosis with some of the histopathological descriptions matching the ground truth): This score is assigned when the report fails to provide the correct diagnosis but includes practical histopathological descriptions that align with the reference report. While some microscopic features are correctly described, the overall diagnostic conclusion is incorrect, greatly reducing the clinical reliability and utility of the report. \\
 - Scores 0 (Incorrect diagnosis with no histopathological descriptions matching with ground truth): This score is assigned to reports that provide neither a correct diagnosis nor any histopathological descriptions that align with the ground truth. These reports fail to recognize key pathological features and do not contribute to an accurate clinical assessment, making them completely unreliable. \\
\\
Ground Truth : \{Ground Truth Report\}  \\
Candidate Report: \{Candidate Report\}  

\end{tcolorbox}

\caption{Prompt templates used for GPT-4.1 analysis.}
\label{fig:gpt4.1prompt}
\end{figure*}

\end{document}